\documentclass[11pt,a4paper]{article}
\usepackage[hyperref]{acl2019}
\usepackage{times}
\usepackage{algorithm}
\usepackage{algorithmic}
\usepackage{footnote}
\makesavenoteenv{algorithm}
\usepackage{amsmath}
\usepackage{dsfont}
\usepackage{mathtools}
\usepackage{booktabs}
\DeclarePairedDelimiter{\floor}{\lfloor}{\rfloor}

\usepackage{latexsym}
\DeclareMathOperator*{\argmax}{argmax}
\usepackage{url}

\RequirePackage[textsize=tiny,backgroundcolor=orange!20,linecolor=red!50]{todonotes}%

\aclfinalcopy % Uncomment this line for the final submission
\def\aclpaperid{1168} %  Enter the acl Paper ID here

\title{FIESTA: Fast IdEntification of State-of-The-Art models using adaptive bandit algorithms}

\author{Henry B. Moss \\
	STOR-i Centre for \\
	Doctoral Training, \\
	Lancaster University \\\And
	Andrew Moore \\
	School of Computing \\and Communications,\\
	Lancaster University \\
	\hspace*{4.5cm}{\tt initial.surname@lancaster.ac.uk} \\\And
	David S. Leslie \\
    Dept. of Mathematics \\
	and Statistics, \\
	Lancaster University \\ \And
	Paul Rayson \\
	School of Computing \\and Communications, \\
	Lancaster University \\	
}

\date{}

\begin{document}
\maketitle

\begin{abstract}
We present FIESTA, a model selection approach that significantly reduces the computational resources required to reliably identify  \textit{state-of-the-art} performance from large collections of candidate models. Despite being known to produce unreliable comparisons, it is still common practice to compare model evaluations based on single choices of random seeds. We show that reliable model selection also requires evaluations based on multiple train-test splits (contrary to common practice in many shared tasks). Using bandit theory from the statistics literature, we are able to adaptively determine appropriate numbers of data splits and random seeds used to evaluate each model, focusing computational resources on the evaluation of promising models whilst avoiding wasting evaluations on models with lower performance. Furthermore, our user-friendly Python implementation produces confidence guarantees of correctly selecting the optimal model. We evaluate our algorithms by selecting between $8$ target-dependent sentiment analysis methods using dramatically fewer model evaluations than current model selection approaches.
\end{abstract}
\section{Introduction and Background}
\label{sec:introduction}

Natural Language Processing (NLP) is a field driven by empirical evaluations. Authors are under pressure to demonstrate that their models or methods achieve \textit{state-of-the-art} performance on a particular task or dataset, which by definition requires reliable model comparison. As models become more numerous, require larger computational resources to train, and the performance of competing models gets closer, the task of reliable model selection has not only become more important, but also increasingly difficult. Without full disclosure of model settings and data splits, it is impossible to accurately compare methods and models.

To be able to perform meaningful model comparisons, we need to be able to reliably evaluate models. Unfortunately, evaluating a model is a non-trivial task and the best we can do is to produce noisy estimates of model performance with the following two distinct sources of stochasticity:
\begin{enumerate}
	\item We only have access to a finite training dataset, however, evaluating a model on its training data leads to severe over-estimates of performance. To evaluate models without over-fitting, practitioners typically randomly partitioning data into independent training and testing sets, producing estimates that are random quantities with often high variability for NLP problems  \cite{Moss2018}. Although methods like bootstrapping \cite{efron1994introduction} and leave-one-out cross validation \cite{kohavi1995study} can provide deterministic estimates of performance, they require the fitting of a large number of models and so are not computationally feasible for the complex models and large data prevalent in NLP. Standard NLP model evaluation strategies range from using a simple (and computationally cheap) single train-test split, to the more sophisticated $K$-fold cross validation, CV \cite{kohavi1995study}.
	\item The vast majority of recent NLP models are non-deterministic and so their performance has another source of stochasticity, controlled by the choice of random seed during training. Common sources of model instability in modern NLP include weight initialisation, data sub-sampling for stochastic gradient calculation, negative sampling used to train word embeddings \cite{mikolov2013distributed} and feature sub-sampling for ensemble methods. In particular, the often \textit{state-of-the-art} LSTMs (and its many variants) have been shown to exhibit high sensitivity to random seeds \cite{reimers2017reporting}.
\end{enumerate}

For reliable model selection, it is crucial to take into account both sources of variability when estimating model performance. Observing a higher score for one model could be a consequence of a particularly non-representative train-test split and/or random seed used to evaluate the model rather than a genuine model improvement. This subtlety is ignored by large scale NLP competitions such as \citet{semeval-2018-international} with evaluations based on a pre-determined train-test split.

Although more precise model evaluations can be obtained with higher computation, calculating overly precise model evaluations is a huge waste of computational resource. On the other hand, our evaluations need to provide reliable conclusions (with only a small probability of selecting a sub-optimal model). It is poorly understood how to choose an appropriate evaluation strategy for a given model selection problem. These are task specific, depending on model stability, the closeness in performance of competing models and subtle properties of the data such as the representativeness of train-test splits.

In contrast to common practice, we consider model selection as a sequential process. Rather than using a fixed evaluation strategy for each model (which we refer to as a non-adaptive approach), we start with a cheap evaluation of each model on just a single train-test split, and then cleverly choose where to allocate further computational resources based on the observed evaluations. If we decide to further test a promising model, we calculate an additional evaluation based on another data split and seed, observing both sources of evaluation variability and allowing reliable assessments of performance. %\footnote{Although the building up of performance estimates using multiple train-test splits does not necessarily produce as large a variance reduction as the more sophisticated $K$-fold CV for the same computational cost, we can easily estimate the variance of an estimator based on train-test splits as the sample variance of the splits. $K$-fold CV's variability is famously difficult to estimate \cite{bengio2004no}, which prevents its use in our adaptive algorithms. Additionally, combining estimators based on train-test splits allows us superior control of computational resources, i.e we can choose to reduce variability by evaluating a single additional train-test splits rather than being forced to fit the $K$ extra models required for a full additional round of $K$-fold CV.} 

To perform sequential model fitting,  we borrow methods from the multi-armed-bandit (MAB) statistical literature  \cite{lai1985asymptotically}. This field covers problems motivated by designing optimal strategies for pulling the arms of a bandit (also known as a slot machine) in casinos. Each arm produces rewards from different random distributions which the user must learn by pulling arms. In particular, model selection is equivalent to the problem of best-arm-identification; identifying the arm with the highest mean. Although appearing simple at a first glance, this problem is deceptively complex and has provided motivation for efficient algorithms in a wide range of domains, including clinical trials \cite{villar2015multi} and recommendation systems \cite{li2010contextual}. 

Although we believe that we are the first to use bandits to reduce the cost and improve the reliability of model selection, we are not the first to use them in NLP. Recent work in machine translation makes use of another major part of the MAB literature, seeking to optimise the long-term performance of translation algorithms \cite{nguyen2017reinforcement,sokolov2016learning,lawrence2017counterfactual}. Within NLP, our work is most similar to \citet{haffari2017efficient}, who use bandits to minimise the number of data queries required to calculate the F-scores of models. However, this work does not consider the stochasticity of the resulting estimates or easily extend to other evaluation metrics.

The main contribution of this paper is the application of three intuitive algorithms to model selection in NLP, alongside a user-friendly Python implementation: FIESTA (Fast IdEntification of State-of-The-Art)\footnote{\url{https://github.com/apmoore1/fiesta}}. We can automatically identify an optimal model from large collections of candidate models to a user-chosen confidence level in a small number of model evaluations. We focus on three distinct scenarios that are of interest to the NLP community. Firstly, we consider the \textbf{fixed budget} (FB) model selection problem (Section \ref{FB}), a situation common in industry, where a fixed quota of computational resources (or time constraints for real-time decisions) must be appropriately allocated to identify an optimal model with the highest possible confidence. In contrast, we also consider the \textbf{fixed confidence} (FC) problem (Section \ref{FC}), which we expect to be of more use for researchers. Here, we wish to claim with a specified confidence level that our selected model is state-of-the-art against a collection of competing models using the minimal amount of computation. Finally, we also consider an extension to the FC scenario, where a practitioner has the computational capacity to fit multiple models in parallel. We demonstrate the effectiveness of our procedures over current model selection approaches when identifying an optimal target-dependent sentiment analysis model from a set of eight competing candidate models (Section \ref{experiments}).

\section{Motivating example}
\begin{figure}[ht]
	\vskip 0.2in
	\begin{center}
		\centerline{\includegraphics[width=\columnwidth]{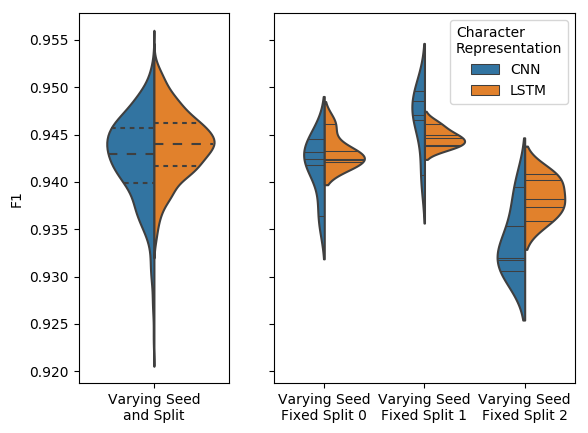}}
		\caption{The left plot shows the distribution of results when varying the data splits and random seeds, with the dashed lines representing the quartile values. The three right plots each represent a different single data split over five runs on different random seeds. The lines represent a single run result.}
		\label{Motivation}
	\end{center}
	\vskip -0.2in
\end{figure}

We now provide evidence for the need to vary both data splits and random seeds for reliable model selection. We extend the motivating example used in the work of \citet{reimers2017reporting}, comparing two LSTM-based Named Entity Recognition (NER) models by \citet{ma2016} and \citet{lample2016}, differing only in character representation (via a CNN and a LSTM respectively). We base model training on \citet{ma2016}, however, following the settings of \citet{yang2018} we use a batch size of 64, a weight decay of $10e^{-9}$ and removed momentum. We ran each of the NER models five times with a different random seed on 150 different train, validation, and test splits\footnote{The original CoNLL data was split with respect to time rather than random sub-sampling, explaining the discrepancy with previous scores on this dataset using the same models.}. \citet{reimers2017reporting} showed the effect of model instability between these two models, where changing the model's random seeds can lead to drawing different conclusions about which model performed best. We extend this argument by showing that different conclusions can also be drawn if we instead vary the train-test split used for the model evaluation (Figure \ref{Motivation}). We see that while data splits 0 and 2 correctly suggest that the LSTM is optimal, using data split 1 suggests the opposite. Therefore, it is clear that we must vary both the random seeds and train-test splits used to evaluate our models if we want reliable model selection.

\section{Problem Statement}
\label{sec:problemstatement}
Extending notation from \citet{arcuri2014hitchhiker}, we can precisely state the task of selecting between a collection of $N$ candidate models $S=\{m_1,m_2,..m_N\}$ as finding
\begin{equation}
    m^*=\argmax_{m\in S}\mathcal{M}(m).
    \label{simpleform}
\end{equation}
$m^*$ is the best model according to some chosen evaluation metric $\mathcal{M}$ that measures the performance of that model, e.g accuracy, F-score or AUC (for an summary of model evaluation metrics see \citet{friedman2001elements}). 

As already argued, Equation (\ref{simpleform}) paints an overly simplistic picture of model selection. In reality we only have access to noisy realisations of the true model score $\mathcal{M}(m)$ and direct comparisons of single realisations of random variables are unreliable. Therefore, we follow the arguments of \citet{reimers2018comparing} and consider a meaningful way of comparing noisy model evaluations: namely, finding the model with largest expected performance estimate across different train-test splits and random seeds. Defining the mean performance of model $m$ as $\mu_m$, we see that the task of model selection is equivalent to the accurate learning and comparison of these $N$ unknown means:
\[
m^*=\argmax_{m\in S}\mu_m.
\]

We can now set up the sequential framework of our model selection procedure and precisely state what we mean by reliable model selection. At each step in our algorithm we choose a model to evaluate and sample a performance estimate by randomly generating a data split and random seed. After collecting evaluations, we can calculate sample means for each model, which we denote as $\Hat{\mu}_m$. After running our algorithm for $T$ steps, reliable model selection corresponds to knowing how confident we should be that our chosen model $\hat{m}_T=\argmax\hat{\mu}_m$ is in fact the true optimal model $m^*$, i.e. we wish to make a precise statement of the form;
\begin{align}
\mathds{P}\left(\hat{m}_T=m^*\right)\geq1-\delta,
\label{Conc}
\end{align} where $1-\delta$ represents this confidence.

In Section \ref{sec:introduction} we motivated two distinct goals of a sequential model selection routine, which we can now state as:
\begin{enumerate}
	\item \textbf{Fixed budget model selection} (FB): We wish to find the best model using only a fixed budget of $T$ model evaluations. The aim is to collect the $T$ evaluations that allow us to claim (\ref{Conc}) with the largest possible confidence level $1-\delta$.
	\item \textbf{Fixed confidence model selection} (FC): We wish to find the best model to a pre-specified confidence level. The aim is to collect the minimal number of model evaluations that allow us to claim (\ref{Conc}).
\end{enumerate}
Although an algorithm designed to do well in one of these scenarios will likely also do well in the other, we will see that to achieve the best performance at either FB or FC model selection, we require subtly different algorithms.

\section{Algorithms}
\label{sec:MAB}
We now examine model selection from a bandit viewpoint, summarising three bandit algorithms and relating their use to three distinct model selection scenarios. Although the underpinning theoretical arguments for these algorithms are beyond the scope of this work, we do highlight one point that is relevant for model selection; that scenarios enjoying  the  largest  efficiency  gains  from moving  to  adaptive  algorithms are  those  where only a subset of arms have performance close to optimal \cite{jamieson2013finding}. Model selection in NLP is often in this scenario,  with only a small number of considered models being close to \textit{state-of-the-art}, and so (as we demonstrate in Section \ref{experiments}) NLP has a lot to gain from using our adaptive model selection algorithms.

\subsection{Fixed Budget by Sequential Halving}
\label{FB}

\begin{algorithm}
\caption{Sequential Halving for Fixed Budget Model Selection}
\label{SeqHalv}
\begin{algorithmic} 
\REQUIRE Computational Budget $T$, \\ \quad \quad \quad \   Set of $N$ candidate models $S$
\WHILE{$|S|\neq1$}
\STATE Evaluate each model $m$ in $S$ $\floor[\Big]{\frac{T}{|S|\lceil\log_2N\rceil}}$ times
\STATE Update the empirical means $\hat{\mu}_m$
\STATE Remove $\floor[\big]{\frac{|S|}{2}}$ models with worst $\hat{\mu}_m$ from $S$
\ENDWHILE
\RETURN Chosen model $S$
\end{algorithmic}
\end{algorithm}

FB best-arm identification algorithms are typically based on successively eliminating arms until just a single (ideally) optimal arm remains \cite{jamieson2013finding,jamieson2014best,audibert2010best}. We focus on the \textbf{sequential halving} (SH) algorithm of \citet{karnin2013almost} (Algorithm \ref{SeqHalv}). Here we break our model selection routine into a series of $\floor[\big]{log_2N}$ rounds, each discarding the least promising half of our candidate model set, eventually resulting in a single remaining model. Our computational budget $T$ is split equally among the rounds to be equally budgeted among the models remaining in that round. This allocation strategy ensures an efficient use of resources, for example the surviving final two models are evaluated $2^{\floor[\big]{log_2N}}-1$ times as often as the models eliminated in the first round. An example run of the algorithm is summarised in Table \ref{exampleseqhalv}.

\begin{table}[ht]
\begin{tabular}{@{}lll@{}}
\toprule
Round & Candidate Models & \# Evaluations \\ \midrule
1 & $S=\{m_1,m_2,m_3,m_4\}$ & 2 \\
2 & $S=\{m_2,m_4\}$ & 4 \\ \midrule
output: & $S=\{m_2\}$ &  \\ \bottomrule
\end{tabular}
\caption{An example of sequential elimination selecting between four models with a budget of $T=16$. After two evaluations of each model, two models are eliminated. The remaining budget is then used to reliably decide between the remaining pair. Standard practice would evaluate each model an equal four times, wasting computational resources on sub-optimal models.} 
\label{exampleseqhalv}
\end{table}

In the bandit literature \cite{karnin2013almost}, this algorithm is shown to have strong theoretical guarantees of reliably choosing the optimal arm, as long as the reward-distributions for each arm are bounded (limited to some finite range). This is not a restrictive assumption for NLP, as the majority of common performance metrics  are bounded, for example accuracy, recall, precision and F-score are all constrained to lie in $\left[0,1\right]$. We will demonstrate the effectiveness of sequential halving for model selection in Section \ref{experiments}.
\subsection{Fixed Confidence by TTTS}
\label{FC}

For fixed confidence model selection, where we wish to guarantee the selection of an optimal arm at a given confidence level, we cannot just discard arms that are likely to be sub-optimal without accurately estimating this likelihood of sub-optimality. Although approaches that sequentially eliminate arms (like our sequential halving algorithm) do exist for FC best-arm identification \cite{jamieson2014lil,karnin2013almost,audibert2010best,even2002pac},  the best theoretical guarantees for the FC problem come from algorithms that maintain the ability to sample any arm at any point in the selection procedure \cite{garivier2016optimal,jamieson2014best}. Rather than seeking to eliminate half the considered models at regular intervals of computation, a model is only evaluated until we can be sufficiently confident that it is sub-optimal. Unfortunately, the performance guarantees for these methods are asymptotic results (in the number of arms and the number of arm pulls) and have little practical relevance to the (at most) tens of arms in a model selection problem.

Our practical recommendation for FC model selection is a variant of the well-known Bayesian sampling algorithm, Thompson sampling, known as \textbf{top-two Thompson sampling} (TTTS) \cite{russo2016simple}. We will see that this algorithm can efficiently allocate computational resources to quickly find optimal models. Furthermore, this approach provides full uncertainty estimation over the final choice of model, providing the confidence guarantees required for FC model selection. 

Our implementation makes the assumption that the evaluations of each model roughly follow a Gaussian distribution, with different means and variances. Although such assumptions  are common in the model evaluation literature \cite{reimers2018comparing} and for statistical testing in NLP \cite{dror2018hitchhiker}, they could be problematic for the bounded metrics common in NLP. Therefore we also experimented with modelling the logit transformation of our evaluations, mapping our evaluation metric to the whole real line. However, for our examples of Section \ref{experiments} we found that this mapping provided a negligible improvement in reliability and so was not worth including in our experimental results. This may not be the case for other tasks or less well-behaved evaluation metrics and so we include this functionality in the FIESTA package.

\begin{algorithm}
\caption{Top-Two Thompson Sampling}
\label{TS}
\begin{algorithmic} 
\REQUIRE Desired Confidence $1-\delta$, \\ \quad \quad \quad \   Set of $N$ candidate models $S$
\STATE Initialise a uniform belief $\pi$
\STATE Evaluate each model in $S$ three times \footnote{We enforce a minimum of three evaluations to ensure that the t distribution in our posterior remains well-defined }
\STATE Update belief $\pi$
\WHILE{$\max_{m\in S}\pi_m\leq 1-\delta$}
\STATE Sample distinct $m_1$ and $m_2$ according to $\pi$
\STATE Randomly choose between $m_1$ and $m_2$
\STATE Evaluate chosen model
\STATE Update belief $\pi$
\ENDWHILE
\RETURN Chosen model $\argmax_{m\in S}\pi_m$
\end{algorithmic}
\end{algorithm}

\begin{figure}[ht]
	\vskip 0.2in
	\begin{center}
		\centerline{\includegraphics[width=\columnwidth]{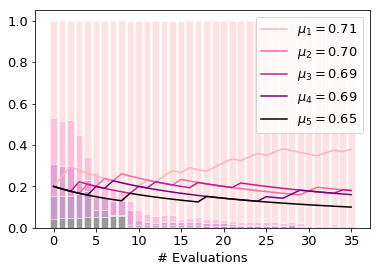}}
		\caption{TTTS seeking the optimal model with confidence $0.99$ from $5$ synthetic models. The background represents our evolving belief $\pi$ in the optimal model and the lines represent the proportion of the total evaluations made on each model. We start evaluating the models uniformly but our adaptive algorithm quickly focuses resources on the best models.}
		\label{TTTS}
	\end{center}
	\vskip -0.2in
\end{figure}

To provide efficient model selection, we use our current believed probability that a given model is optimal $\pi_m=\mathds{P}\left(m^*=m\right)$ (producing a distribution over the models $\pi=\{\pi_1,..,\pi_N\}$) to drive the allocation of computational resources. Standard Thompson sampling is a stochastic algorithm that generates a choice of model by sampling from our current belief $\pi$, i.e. choosing to evaluate a model with the same probability that we believe is optimal (see \citet{russo2018tutorial} for a concise introduction). Although this strategy allows us to focus computation on promising arms, it actually does so too aggressively. Once we believe that an arm is optimal with reasonably high confidence, computation will be heavily focused on evaluating this arm even though we need to become more confident about the sub-optimality of competing models to improve our confidence level. This criticism motivates our chosen algorithm TTTS (summarised in Algorithm \ref{TS}), where instead of sampling a single model according to $\pi$, we sample two distinct models. We then uniformly choose between these two models for the next evaluation, allowing a greater exploration of the arms and much improved rates of convergence to the desired confidence level \cite{russo2016simple}. We use this new evaluation to update our belief and continue making evaluations until we believe that a model is optimal with a higher probability than $1-\delta$ and terminate the algorithm. An example run of TTTS is demonstrated on a synthetic example in Figure \ref{TTTS}, where we simulate from $5$ Gaussian distributions with means $\{0.65,0.69,0.69,0.70,0.71\}$ and standard deviation $0.01$ to mimic accuracy measurements for a model selection problem.

We now explain how we calculate $\pi$ (our belief in the location of the optimal model) using well-known results from Bayesian decision theory (see \citet{berger2013statistical} for a comprehensive coverage). As justified earlier, we assume that the evaluations of model $m$ are independently distributed with a Gaussian distribution $\mathcal{N}(\mu_m,\sigma_m^2)$ for some unknown mean $\mu_m$ and variance $\sigma_m^2$. Although we are primarily interested in learning $\mu_m$, we must also learn $\sigma_m^2$ in order to make confidence guarantees about the optimality of our selected model. Therefore, as well as keeping track of the sample means for the evaluations of each model $\hat{\mu}_m$, we also keep track of the sample variances $\hat{S}_m$ and counters $T_m$ of the number of times each model has been evaluated. To facilitate inference, we choose a uniform prior for the unknown $\mu_m$ and $\sigma_m$. Not only is this a conjugate prior for Gaussian likelihoods, but it is also shown to encourage beneficial exploratory behaviour when using Thompson sampling on Gaussian bandit problems \cite{honda2014optimality} and so allows fast identification of optimal arms (or models). After observing $T_m$ evaluations of each model and producing estimates $\hat{\mu}_m$ and $\Hat{S}_m$,  our posterior belief for each deviation between the true and observed model means $\mu_m-\hat{\mu}_m$ satisfies (as derived in \cite{honda2014optimality});
\[
\sqrt{\frac{T_m(T_m-2)}{\hat{S}_m}}\left(\mu_m-\Hat{\mu}_m\right)|\,\hat{\mu}_m,\hat{S}_m\sim t_{T_m-2}, 
\]
where $t_d$ is a Student's t-distribution with $d$ degrees of freedom. 

$\pi$ is then defined as the probability vector, such that $\pi_m$ is the relative probability that $\mu_m$ is the largest according to this posterior belief. Unfortunately, there is no closed form expression for the maximum of $N$ t-distributions and so FIESTA uses a simple Monte-Carlo approximation based on the sample maxima of repeated draws from our posteriors for $\mu_m$. In practice this is very accurate and did not slow down our experiments, especially in comparison to the time saved by reducing the number of model evaluations.

\subsection{Batch Fixed Confidence by BTS}

NLP practitioners often have the computational capacity to fit models in parallel across multiple workers, evaluating multiple models or the same model across multiple seeds at once. Their model selection routines must therefore provide batches of models to evaluate. Our proposed solution to FB model selection naturally provides such batches, with each successive round of SH producing a collection of model evaluations that can be calculated in parallel. Unfortunately, TTTS for FC model selection successively chooses and then waits for the evaluation of single models and so is not naturally suited to parallelism. 

Extending TTTS to batch decision making is an open problem in the MAB literature. Therefore, we instead consider \textbf{batch Thompson sampling} (BTS), an extension of standard Thompson sampling (as described in Section \ref{FC}) to batch sampling from the related field of Bayesian optimisation \cite{kandasamy2018parallelised}. At each step in our selection process we take $B$ model draws according to our current belief $\pi$ that the model is optimal, where $B$ represents our computational capacity. This is in contrast to the single draw in standard Thompson sampling and the drawn pair in TTTS. In addition, this approach extends to the asynchronous setting, where rather than waiting for the whole batch of $B$ models to be evaluated before choosing the next batch, each worker can draw a new model to evaluate according to the updated $\pi$. This flexibility provides an additional efficiency gain for problems where the different models have a wide range of run times.

\section{Experiments}
\label{experiments}
We now test our three algorithms on a challenging model selection task typical of NLP, selecting between eight Target Dependent Sentiment Analysis (TDSA) models based on their macro F1 score. We consider two variants of four re-implementations of well-known TDSA models: ATAE \cite{wang2016}, IAN \cite{ma2017}, TDLSTM \cite{tang2016} (without target words in the left and right LSTM), and a non-target-aware LSTM method used as the baseline in \citet{tang2016}. 

\begin{figure}[ht]
	\vskip 0.2in
	\begin{center}
		\centerline{\includegraphics[width=\columnwidth]{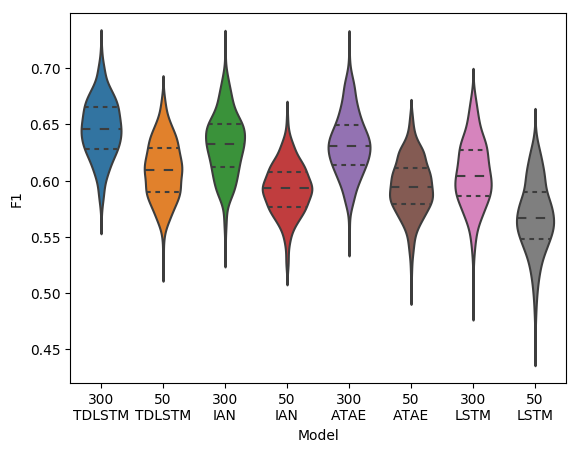}}
		\caption{F1 scores for our candidate TDSA models. After $500$ evaluations of each model on different data splits and model seeds we see that the TDLSTM is the \textit{state-of-the-art} model. 
		}
		\label{TDSA}
	\end{center}
	\vskip -0.2in
\end{figure}

These methods represent \textit{state-of-the-art} within TDSA, with only small differences in performance between TDLSTM, IAN, and ATAE (see figure \ref{TDSA}). All the models are re-implemented in PyTorch \cite{paszke2017} using AllenNLP \cite{gardner2018}. To ensure the only difference between the models is their network architecture the models use the same optimiser settings and the same regularisation.  All words are lower cased and we use the same Glove common crawl 840B token 300 dimension word embedding \cite{pennington2014}. We use variational \cite{gal2016} and regular \cite{hinton2012} dropout for regularisation and an ADAM \cite{kingma2014} optimiser with standard settings, a batch size of $32$ and use at most $100$ epochs (with early stopping on a validation set). Many of these settings are not the same as originally implemented, however, having the same training setup is required for fair comparison (this explains the differences between our results and the original implementations). To increase the difficulty of our  model selection problem, we additionally create four extra models by reducing the dimensions of the Glove vectors to 50 and removing dropout. Although these models are clearly not \textit{state-of-the-art}, they increase the size of our candidate model set and so provide a more complicated model selection problem (an intuition discussed in Appendix \ref{TheoreticalProperties}).

All of the TDSA experiments are conducted on the well-studied SemEval 2014 task 4 Restaurant dataset \cite{pontiki2014} and we force train-val-test splits to follow the same ratios as this dataset's official train-test split.  Each individual model evaluation is then made on a randomly generated train-test split and random seed to access both sources of evaluation variability.

\subsection{Fixed Budget Model Selection}
We use the TDSA model selection problem to test fixed budget model selection. To thoroughly test our algorithm, we consider an additional four models based on 200 dimensional Glove vectors, bringing the total number of models to 12. We compare our approach of sequential halving to the standard non-adaptive approach of splitting the available computational budget equally between the 12 candidate models. For example, we would allocate a budget of $24$ model evaluations as evaluating each model two times and selecting the model with the highest sample mean.

\begin{figure}[ht]
	\vskip 0.2in
	\begin{center}
		\centerline{\includegraphics[width=\columnwidth]{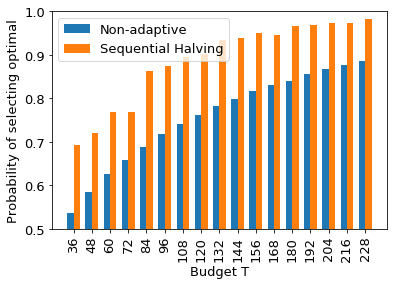}}
		\caption{Proportion of the runs correctly selecting the optimal TDSA model using sequential halving against the standard non-adaptive approach. Sequential halving consistently identifies the optimal model at a significantly higher rate across  a wide range of budgets.}
		\label{SHfig}
	\end{center}
	\vskip -0.2in
\end{figure}

Figure \ref{SHfig} compares the proportion of $10,000$ runs of sequential halving that correctly identify the optimal model with the proportion identified by the non-adaptive approach with the same computational budget. Sequential halving identifies the optimal model more reliably ($\approx15\%$ more often) than the current approach to FB model selection in NLP. Using sequential halving with $204$ evaluations almost always ($99\%$ of runs) selects the optimal model, whereas the non-adaptive approach is only correct $85\%$ of the time.

\begin{table*}[ht!]
\centering
\begin{tabular}{l|llll|llll}
\hline
 & \# & evaluations & with & Non-Adaptive & \# & evaluations & with & TTTS \\ \hline
$\delta$ & \multicolumn{1}{l|}{min} & \multicolumn{1}{l|}{mean} & \multicolumn{1}{l|}{max} & \begin{tabular}[c]{@{}l@{}}$\%$ correctly\\ selected\end{tabular} & \multicolumn{1}{l|}{min} & \multicolumn{1}{l|}{mean} & \multicolumn{1}{l|}{max} & \begin{tabular}[c]{@{}l@{}}$\%$ correctly\\ selected\end{tabular} \\ \hline
0.05 & 48 & 281 & 1552 & 100 & 27 & 130 & 518 & 100 \\
0.1 & 40 & 206 & 1192 & 99 & 24 & 96 & 460 & 99 \\
0.2 & 32 & 128 & 608 & 96 & 24 & 65 & 274 & 97 \\ \hline
\end{tabular}
\caption{Number of evaluations required to select a TDSA model at a range of confidence levels across $500$ runs of TTTS and a standard non-adaptive approach.}
\label{TTTStable}
\end{table*}

\subsection{Fixed Confidence Model Selection}
We perform fixed confidence model selection on the eight TDSA candidate models (the full models and those based on 50 dimensional vectors). We compare TTTS to a non-adaptive approach where all models are evaluated at each step, irrespective of the results of earlier evaluations (the standard approach for model selection in NLP). We run this non-adaptive approach until we reach the required confidence level calculated using the same Bayesian framework as in TTTS.

We run each approach $500$ times and note the number evaluations required to get to a range of confidence levels (Table \ref{TTTStable}) alongside the proportion that correctly identify the optimal model. TTTS requires substantially less model evaluations (in terms of the minimum, mean and max across our runs) to reach a given confidence level than the non-adaptive approach, achieving the same reliability at half the cost (on average). TTTS is able to quickly identify sub-optimal models and so can avoid wasting resources repeatedly evaluating the whole candidate set.
\begin{table*}[ht!]
\centering
\begin{tabular}{l|llll|llll}
\hline
 & \# & evaluations & with & BTS-4 & \# & evaluations & with & BTS-8 \\ \hline
$\delta$ & \multicolumn{1}{l|}{min} & \multicolumn{1}{l|}{mean} & \multicolumn{1}{l|}{max} & \begin{tabular}[c]{@{}l@{}}$\%$ correctly\\ selected\end{tabular} & \multicolumn{1}{l|}{min} & \multicolumn{1}{l|}{mean} & \multicolumn{1}{l|}{max} & \begin{tabular}[c]{@{}l@{}}$\%$ correctly\\ selected\end{tabular} \\ \hline
0.05 & 28 & 282 & 1392 & 100 & 88 & 315 & 1128 & 100 \\
0.1 & 24 & 144 & 520 & 100 & 56 & 178 & 784 & 100 \\
0.2 & 24 & 76 & 280 & 98 & 32 & 106 & 352 & 99 \\ \hline
\end{tabular}
\caption{Number of evaluations of required to select a TDSA model at a range of confidence levels across $500$ runs of BTS selecting batches of 4 and 8 models.}
\label{BTStable}
\end{table*}

Finally, we test our proposed approach to batch FC model selection by running exactly the same experiment but using BTS to choose collections of four and eight models at a time (Table \ref{BTStable}). As expected, performance degrades as we increase batch size, with batches of four allowing more fine grained control over model evaluations than using batches of eight. In particular, due to the exploitative nature of Thompson sampling, we see that selecting models to a very high confidence (95\%) requires more computation with BTS than the standard non-adaptive approach. However, BTS does reach the other confidence levels faster and correctly identifies the optimal model more often. However, as TTTS performs significantly better across all confidence levels, we emphasise the need for a less-exploitative version of BTS with adjustments similar to those used in TTTS.

\section{Conclusions}

The aim of this paper has been to propose three algorithms for model selection in NLP, providing efficient and reliable selection for two distinct realistic scenarios: fixed confidence and fixed budget model selection. Crucially, our research further calls into question the current practice in NLP evaluation as used in the literature and international competitions such as SemEval. Our algorithms adaptively allocate resources to evaluate promising models, basing evaluations across multiple random seeds and train-test splits. We demonstrate that this allows significant computational savings and improves reliability over current model selection approaches.

Although we have demonstrated that our algorithms perform well on a complex model selection problem typical of NLP, there is still work to be done to create algorithms more suited to these problems. Future research directions include making selection routines more robust to evaluation outliers, relaxing our Gaussian assumptions and developing more effective batch strategies.

\section{Acknowledgements}
The authors are grateful to reviewers, whose comments and advice have greatly improved this paper. The research was supported by an EPSRC Doctoral Training Grant and the STOR-i Centre for Doctoral Training. We thank Dr Chris Jewell at the Centre for Health Informatics, Computing, and Statistics, Lancaster University for the loan of a NVIDIA GP100-equipped workstation for this study.

\bibliography{acl2019}

\begin{thebibliography}{43}
\expandafter\ifx\csname natexlab\endcsname\relax\def\natexlab#1{#1}\fi

\bibitem[{Arcuri and Briand(2014)}]{arcuri2014hitchhiker}
Andrea Arcuri and Lionel Briand. 2014.
\newblock \href {https://doi.org/10.1002/stvr.1486} {A hitchhiker's guide to
  statistical tests for assessing randomized algorithms in software
  engineering}.
\newblock \emph{Software Testing, Verification and Reliability},
  24(3):219--250.

\bibitem[{Audibert and Bubeck(2010)}]{audibert2010best}
Jean-Yves Audibert and S{\'e}bastien Bubeck. 2010.
\newblock \href
  {https://hal-enpc.archives-ouvertes.fr/hal-00654404/file/COLT10.pdf} {Best
  arm identification in multi-armed bandits}.
\newblock In \emph{COLT - 23th Conference on Learning Theory - 2010}, pages
  13--p.

\bibitem[{Berger(2013)}]{berger2013statistical}
James~O Berger. 2013.
\newblock \emph{Statistical decision theory and Bayesian analysis}.
\newblock Springer Science \& Business Media.

\bibitem[{Dror et~al.(2018)Dror, Baumer, Shlomov, and
  Reichart}]{dror2018hitchhiker}
Rotem Dror, Gili Baumer, Segev Shlomov, and Roi Reichart. 2018.
\newblock \href {http://aclweb.org/anthology/P18-1128} {The hitchhiker's guide
  to testing statistical significance in natural language processing}.
\newblock In \emph{Proceedings of the 56th Annual Meeting of the Association
  for Computational Linguistics (Volume 1: Long Papers)}, pages 1383--1392.
  Association for Computational Linguistics.

\bibitem[{Efron and Tibshirani(1994)}]{efron1994introduction}
Bradley Efron and Robert~J Tibshirani. 1994.
\newblock \emph{An introduction to the bootstrap}.
\newblock CRC press.

\bibitem[{Even-Dar et~al.(2002)Even-Dar, Mannor, and Mansour}]{even2002pac}
Eyal Even-Dar, Shie Mannor, and Yishay Mansour. 2002.
\newblock Pac bounds for multi-armed bandit and markov decision processes.
\newblock In \emph{International Conference on Computational Learning Theory},
  pages 255--270. Springer.

\bibitem[{Friedman et~al.(2001)Friedman, Hastie, and
  Tibshirani}]{friedman2001elements}
Jerome Friedman, Trevor Hastie, and Robert Tibshirani. 2001.
\newblock \emph{The elements of statistical learning}.
\newblock Springer series in statistics New York, NY, USA:.

\bibitem[{Gal and Ghahramani(2016)}]{gal2016}
Yarin Gal and Zoubin Ghahramani. 2016.
\newblock \href
  {http://papers.nips.cc/paper/6241-a-theoretically-grounded-application-of-dropout-in-recurrent-neural-networks.pdf}
  {A theoretically grounded application of dropout in recurrent neural
  networks}.
\newblock In D.~D. Lee, M.~Sugiyama, U.~V. Luxburg, I.~Guyon, and R.~Garnett,
  editors, \emph{Advances in Neural Information Processing Systems 29}, pages
  1019--1027. Curran Associates, Inc.

\bibitem[{Gardner et~al.(2018)Gardner, Grus, Neumann, Tafjord, Dasigi, F.~Liu,
  Peters, Schmitz, and Zettlemoyer}]{gardner2018}
Matt Gardner, Joel Grus, Mark Neumann, Oyvind Tafjord, Pradeep Dasigi, Nelson
  F.~Liu, Matthew Peters, Michael Schmitz, and Luke Zettlemoyer. 2018.
\newblock \href {http://aclweb.org/anthology/W18-2501} {Allennlp: A deep
  semantic natural language processing platform}.
\newblock In \emph{Proceedings of Workshop for NLP Open Source Software
  (NLP-OSS)}, pages 1--6. Association for Computational Linguistics.

\bibitem[{Garivier and Kaufmann(2016)}]{garivier2016optimal}
Aur{\'e}lien Garivier and Emilie Kaufmann. 2016.
\newblock \href {http://proceedings.mlr.press/v49/garivier16a.pdf} {Optimal
  best arm identification with fixed confidence}.
\newblock In \emph{Conference on Learning Theory}, pages 998--1027.

\bibitem[{Haffari et~al.(2017)Haffari, Tran, and Carman}]{haffari2017efficient}
Gholamreza Haffari, Tuan~Dung Tran, and Mark Carman. 2017.
\newblock \href {http://aclweb.org/anthology/E17-1039} {Efficient benchmarking
  of nlp apis using multi-armed bandits}.
\newblock In \emph{Proceedings of the 15th Conference of the European Chapter
  of the Association for Computational Linguistics: Volume 1, Long Papers},
  pages 408--416. Association for Computational Linguistics.

\bibitem[{Hinton et~al.(2012)Hinton, Srivastava, Krizhevsky, Sutskever, and
  Salakhutdinov}]{hinton2012}
Geoffrey~E Hinton, Nitish Srivastava, Alex Krizhevsky, Ilya Sutskever, and
  Ruslan~R Salakhutdinov. 2012.
\newblock \href {https://arxiv.org/pdf/1207.0580.pdf} {Improving neural
  networks by preventing co-adaptation of feature detectors}.
\newblock \emph{arXiv preprint arXiv:1207.0580}.

\bibitem[{Honda and Takemura(2014)}]{honda2014optimality}
Junya Honda and Akimichi Takemura. 2014.
\newblock \href {http://proceedings.mlr.press/v33/honda14.pdf} {Optimality of
  thompson sampling for gaussian bandits depends on priors}.
\newblock In \emph{Artificial Intelligence and Statistics}, pages 375--383.

\bibitem[{Jamieson et~al.(2013)Jamieson, Malloy, Nowak, and
  Bubeck}]{jamieson2013finding}
Kevin Jamieson, Matthew Malloy, Robert Nowak, and Sebastien Bubeck. 2013.
\newblock \href {https://arxiv.org/pdf/1306.3917.pdf} {On finding the largest
  mean among many}.
\newblock \emph{arXiv preprint arXiv:1306.3917}.

\bibitem[{Jamieson et~al.(2014)Jamieson, Malloy, Nowak, and
  Bubeck}]{jamieson2014lil}
Kevin Jamieson, Matthew Malloy, Robert Nowak, and S{\'e}bastien Bubeck. 2014.
\newblock \href {http://proceedings.mlr.press/v35/jamieson14.pdf} {lil’ucb:
  An optimal exploration algorithm for multi-armed bandits}.
\newblock In \emph{Conference on Learning Theory}, pages 423--439.

\bibitem[{Jamieson and Nowak(2014)}]{jamieson2014best}
Kevin Jamieson and Robert Nowak. 2014.
\newblock \href {https://ieeexplore.ieee.org/stamp/stamp.jsp?arnumber=6814096}
  {Best-arm identification algorithms for multi-armed bandits in the fixed
  confidence setting}.
\newblock In \emph{2014 48th Annual Conference on Information Sciences and
  Systems (CISS)}, pages 1--6. IEEE.

\bibitem[{Kandasamy et~al.(2018)Kandasamy, Krishnamurthy, Schneider, and
  P{\'o}czos}]{kandasamy2018parallelised}
Kirthevasan Kandasamy, Akshay Krishnamurthy, Jeff Schneider, and Barnab{\'a}s
  P{\'o}czos. 2018.
\newblock \href
  {http://proceedings.mlr.press/v84/kandasamy18a/kandasamy18a.pdf}
  {Parallelised bayesian optimisation via thompson sampling}.
\newblock In \emph{International Conference on Artificial Intelligence and
  Statistics}.

\bibitem[{Karnin et~al.(2013)Karnin, Koren, and Somekh}]{karnin2013almost}
Zohar Karnin, Tomer Koren, and Oren Somekh. 2013.
\newblock \href {http://proceedings.mlr.press/v28/karnin13.pdf} {Almost optimal
  exploration in multi-armed bandits}.
\newblock In \emph{International Conference on Machine Learning}, pages
  1238--1246.

\bibitem[{Kingma and Ba(2014)}]{kingma2014}
Diederik~P Kingma and Jimmy Ba. 2014.
\newblock \href {https://arxiv.org/pdf/1412.6980.pdf} {Adam: A method for
  stochastic optimization}.
\newblock \emph{arXiv preprint arXiv:1412.6980}.

\bibitem[{Kohavi(1995)}]{kohavi1995study}
Ron Kohavi. 1995.
\newblock \href {https://www.ijcai.org/Proceedings/95-2/Papers/016.pdf} {A
  study of cross-validation and bootstrap for accuracy estimation and model
  selection}.
\newblock In \emph{Proceedings of the 14th international joint conference on
  Artificial intelligence-Volume 2}, pages 1137--1143. Morgan Kaufmann
  Publishers Inc.

\bibitem[{Lai and Robbins(1985)}]{lai1985asymptotically}
Tze~Leung Lai and Herbert Robbins. 1985.
\newblock \href {https://core.ac.uk/download/pdf/82425825.pdf} {Asymptotically
  efficient adaptive allocation rules}.
\newblock \emph{Advances in applied mathematics}, 6(1):4--22.

\bibitem[{Lample et~al.(2016)Lample, Ballesteros, Subramanian, Kawakami, and
  Dyer}]{lample2016}
Guillaume Lample, Miguel Ballesteros, Sandeep Subramanian, Kazuya Kawakami, and
  Chris Dyer. 2016.
\newblock \href {https://doi.org/10.18653/v1/N16-1030} {Neural architectures
  for named entity recognition}.
\newblock In \emph{Proceedings of the 2016 Conference of the North American
  Chapter of the Association for Computational Linguistics: Human Language
  Technologies}, pages 260--270. Association for Computational Linguistics.

\bibitem[{Lawrence et~al.(2017)Lawrence, Sokolov, and
  Riezler}]{lawrence2017counterfactual}
Carolin Lawrence, Artem Sokolov, and Stefan Riezler. 2017.
\newblock \href {https://doi.org/10.18653/v1/D17-1272} {Counterfactual learning
  from bandit feedback under deterministic logging : A case study in
  statistical machine translation}.
\newblock In \emph{Proceedings of the 2017 Conference on Empirical Methods in
  Natural Language Processing}, pages 2566--2576. Association for Computational
  Linguistics.

\bibitem[{Li et~al.(2010)Li, Chu, Langford, and Schapire}]{li2010contextual}
Lihong Li, Wei Chu, John Langford, and Robert~E Schapire. 2010.
\newblock \href
  {https://www.wwwconference.org/proceedings/www2010/www/p661.pdf} {A
  contextual-bandit approach to personalized news article recommendation}.
\newblock In \emph{Proceedings of the 19th international conference on World
  wide web}, pages 661--670. ACM.

\bibitem[{Ma et~al.(2017)Ma, Li, Zhang, and Wang}]{ma2017}
Dehong Ma, Sujian Li, Xiaodong Zhang, and Houfeng Wang. 2017.
\newblock \href {https://www.ijcai.org/proceedings/2017/0568.pdf} {Interactive
  attention networks for aspect-level sentiment classification}.
\newblock In \emph{Proceedings of the 26th International Joint Conference on
  Artificial Intelligence}, pages 4068--4074. AAAI Press.

\bibitem[{Ma and Hovy(2016)}]{ma2016}
Xuezhe Ma and Eduard Hovy. 2016.
\newblock \href {https://doi.org/10.18653/v1/P16-1101} {End-to-end sequence
  labeling via bi-directional lstm-cnns-crf}.
\newblock In \emph{Proceedings of the 54th Annual Meeting of the Association
  for Computational Linguistics (Volume 1: Long Papers)}, pages 1064--1074.
  Association for Computational Linguistics.

\bibitem[{Mannor and Tsitsiklis(2004)}]{mannor2004sample}
Shie Mannor and John~N Tsitsiklis. 2004.
\newblock \href {http://www.jmlr.org/papers/volume5/mannor04b/mannor04b.pdf}
  {The sample complexity of exploration in the multi-armed bandit problem}.
\newblock \emph{Journal of Machine Learning Research}, 5(Jun):623--648.

\bibitem[{Mikolov et~al.(2013)Mikolov, Sutskever, Chen, Corrado, and
  Dean}]{mikolov2013distributed}
Tomas Mikolov, Ilya Sutskever, Kai Chen, Greg~S Corrado, and Jeff Dean. 2013.
\newblock \href
  {https://papers.nips.cc/paper/5021-distributed-representations-of-words-and-phrases-and-their-compositionality.pdf}
  {Distributed representations of words and phrases and their
  compositionality}.
\newblock In \emph{Advances in neural information processing systems}, pages
  3111--3119.

\bibitem[{Moss et~al.(2018)Moss, Leslie, and Rayson}]{Moss2018}
Henry Moss, David Leslie, and Paul Rayson. 2018.
\newblock \href {http://aclweb.org/anthology/C18-1252} {Using j-k-fold cross
  validation to reduce variance when tuning nlp models}.
\newblock In \emph{Proceedings of the 27th International Conference on
  Computational Linguistics}, pages 2978--2989. Association for Computational
  Linguistics.

\bibitem[{Nguyen et~al.(2017)Nguyen, Daum{\'e}~III, and
  Boyd-Graber}]{nguyen2017reinforcement}
Khanh Nguyen, Hal Daum{\'e}~III, and Jordan Boyd-Graber. 2017.
\newblock \href {https://doi.org/10.18653/v1/D17-1153} {Reinforcement learning
  for bandit neural machine translation with simulated human feedback}.
\newblock In \emph{Proceedings of the 2017 Conference on Empirical Methods in
  Natural Language Processing}, pages 1464--1474. Association for Computational
  Linguistics.

\bibitem[{Paszke et~al.(2017)Paszke, Gross, Chintala, Chanan, Yang, DeVito,
  Lin, Desmaison, Antiga, and Lerer}]{paszke2017}
Adam Paszke, Sam Gross, Soumith Chintala, Gregory Chanan, Edward Yang, Zachary
  DeVito, Zeming Lin, Alban Desmaison, Luca Antiga, and Adam Lerer. 2017.
\newblock \href {https://openreview.net/pdf?id=BJJsrmfCZ} {Automatic
  differentiation in pytorch}.
\newblock In \emph{NIPS-W}.

\bibitem[{Pennington et~al.(2014)Pennington, Socher, and
  Manning}]{pennington2014}
Jeffrey Pennington, Richard Socher, and Christopher Manning. 2014.
\newblock \href {https://doi.org/10.3115/v1/D14-1162} {Glove: Global vectors
  for word representation}.
\newblock In \emph{Proceedings of the 2014 Conference on Empirical Methods in
  Natural Language Processing (EMNLP)}, pages 1532--1543. Association for
  Computational Linguistics.

\bibitem[{Pontiki et~al.(2014)Pontiki, Galanis, Pavlopoulos, Papageorgiou,
  Androutsopoulos, and Manandhar}]{pontiki2014}
Maria Pontiki, Dimitris Galanis, John Pavlopoulos, Harris Papageorgiou, Ion
  Androutsopoulos, and Suresh Manandhar. 2014.
\newblock \href {https://doi.org/10.3115/v1/S14-2004} {Semeval-2014 task 4:
  Aspect based sentiment analysis}.
\newblock In \emph{Proceedings of the 8th International Workshop on Semantic
  Evaluation (SemEval 2014)}, pages 27--35. Association for Computational
  Linguistics.

\bibitem[{Reimers and Gurevych(2017)}]{reimers2017reporting}
Nils Reimers and Iryna Gurevych. 2017.
\newblock \href {https://doi.org/10.18653/v1/D17-1035} {Reporting score
  distributions makes a difference: Performance study of lstm-networks for
  sequence tagging}.
\newblock In \emph{Proceedings of the 2017 Conference on Empirical Methods in
  Natural Language Processing}, pages 338--348. Association for Computational
  Linguistics.

\bibitem[{Reimers and Gurevych(2018)}]{reimers2018comparing}
Nils Reimers and Iryna Gurevych. 2018.
\newblock \href {https://arxiv.org/pdf/1803.09578.pdf} {Why comparing single
  performance scores does not allow to draw conclusions about machine learning
  approaches}.
\newblock \emph{arXiv preprint arXiv:1803.09578}.

\bibitem[{Russo(2016)}]{russo2016simple}
Daniel Russo. 2016.
\newblock \href {http://proceedings.mlr.press/v49/russo16.pdf} {Simple bayesian
  algorithms for best arm identification}.
\newblock In \emph{Conference on Learning Theory}, pages 1417--1418.

\bibitem[{Russo et~al.(2018)Russo, Van~Roy, Kazerouni, Osband, Wen
  et~al.}]{russo2018tutorial}
Daniel~J Russo, Benjamin Van~Roy, Abbas Kazerouni, Ian Osband, Zheng Wen,
  et~al. 2018.
\newblock \href {https://djrusso.github.io/RLCourse/papers/TS_Tutorial.pdf} {A
  tutorial on thompson sampling}.
\newblock \emph{Foundations and Trends in Machine Learning}, 11(1):1--96.

\bibitem[{SemEval()}]{semeval-2018-international}
SemEval. 2018.
\newblock \href {https://doi.org/10.18653/v1/S18-1} {\emph{Proceedings of The
  12th International Workshop on Semantic Evaluation}}. Association for
  Computational Linguistics, New Orleans, Louisiana.

\bibitem[{Sokolov et~al.(2016)Sokolov, Kreutzer, Lo, and
  Riezler}]{sokolov2016learning}
Artem Sokolov, Julia Kreutzer, Christopher Lo, and Stefan Riezler. 2016.
\newblock \href {https://doi.org/10.18653/v1/P16-1152} {Learning structured
  predictors from bandit feedback for interactive nlp}.
\newblock In \emph{Proceedings of the 54th Annual Meeting of the Association
  for Computational Linguistics (Volume 1: Long Papers)}, pages 1610--1620.
  Association for Computational Linguistics.

\bibitem[{Tang et~al.(2016)Tang, Qin, Feng, and Liu}]{tang2016}
Duyu Tang, Bing Qin, Xiaocheng Feng, and Ting Liu. 2016.
\newblock \href {http://aclweb.org/anthology/C16-1311} {Effective lstms for
  target-dependent sentiment classification}.
\newblock In \emph{Proceedings of COLING 2016, the 26th International
  Conference on Computational Linguistics: Technical Papers}, pages 3298--3307.
  The COLING 2016 Organizing Committee.

\bibitem[{Villar et~al.(2015)Villar, Bowden, and Wason}]{villar2015multi}
Sof{\'\i}a~S Villar, Jack Bowden, and James Wason. 2015.
\newblock \href {https://doi.org/10.1214/14-STS504} {Multi-armed bandit models
  for the optimal design of clinical trials: benefits and challenges}.
\newblock \emph{Statistical science: a review journal of the Institute of
  Mathematical Statistics}, 30(2):199.

\bibitem[{Wang et~al.(2016)Wang, Huang, zhu, and Zhao}]{wang2016}
Yequan Wang, Minlie Huang, xiaoyan zhu, and Li~Zhao. 2016.
\newblock \href {https://doi.org/10.18653/v1/D16-1058} {Attention-based lstm
  for aspect-level sentiment classification}.
\newblock In \emph{Proceedings of the 2016 Conference on Empirical Methods in
  Natural Language Processing}, pages 606--615. Association for Computational
  Linguistics.

\bibitem[{Yang et~al.(2018)Yang, Liang, and Zhang}]{yang2018}
Jie Yang, Shuailong Liang, and Yue Zhang. 2018.
\newblock \href {http://aclweb.org/anthology/C18-1327} {Design challenges and
  misconceptions in neural sequence labeling}.
\newblock In \emph{Proceedings of the 27th International Conference on
  Computational Linguistics}, pages 3879--3889. Association for Computational
  Linguistics.

\end{thebibliography}
\bibliographystyle{acl_natbib}

\section{Appendix}
\appendix
\section{Characterising the Difficulty of a Model Selection Problem}
\label{TheoreticalProperties}
We briefly summarise a result from the best-arm identification literature, providing intuition for our experiment section through a mechanism to characterise the difficulty of a model selection problem.  Intuitively, model selection difficulty increases with the size of the set of candidate models $N$ and as the performance of sub-optimal models approaches that of the optimal model (and becomes harder to distinguish), i.e. as $\mu_{m^*}-\mu_{m}$ gets small for some sub-optimal arm $m$. In fact, it is well known in the MAB literature that it is exactly these two properties that characterise the complexity of a best-arm-identification problem, confirming our intuition for model selection. \citet{mannor2004sample} show that the number of arm pulls required for the identification of a best arm at a confidence level $1-\delta$ has at least a computational complexity of $O(H\log(1/\delta))$, where
\[
H=\sum_{m' \in S \setminus\{m*\}}\frac{1}{\left(\mu_{m^*}-\mu_{m}\right)^2}.
\]

\end{document}